\documentclass[letterpaper, 10 pt, conference]{ieeeconf}  % Comment this line out if you need a4paper

\IEEEoverridecommandlockouts                              % This command is only needed if 
\usepackage{amsmath} % assumes amsmath package installed
\usepackage{amssymb}  % assumes amsmath package installed
\usepackage{graphicx}
\usepackage{graphbox}
\usepackage{cite}
\usepackage{algorithm}
\usepackage{color}
\usepackage[noend]{algpseudocode}
\usepackage{url}

\newcommand{\approachname}{MSC-VO}
\newcommand{\approachnamelong}{Manhattan and Structural Constraints - Visual Odometry}

\title{\LARGE \bf \approachname: Exploiting Manhattan and \\Structural Constraints for Visual Odometry
}

\author{Joan P. Company-Corcoles, Emilio Garcia-Fidalgo and Alberto Ortiz% <-this % stops a space
\thanks{This work is partially supported by EU-H2020 projects BUGWRIGHT2 (GA 871260) and ROBINS (GA 779776), and by project PGC2018-095709-B-C21 (funded by MCIU/AEI/10.13039/501100011033 and FEDER ``Una manera de hacer Europa"). This publication reflects only the authors views and the European Union is not liable for any use that may be made of the information contained therein.}
\thanks{All authors are with the Department of Mathematics and Computer Science (University of the Balearic Islands) and IDISBA (Institut d'Investigacio Sanitaria de les Illes Balears), Palma de Mallorca, Spain. {\small \{joanpep.company, emilio.garcia, alberto.ortiz\}@uib.es}.}%%
}

\begin{document}

\maketitle
\thispagestyle{empty}
\pagestyle{empty}

%%%%%%%%%%%%%%%%%%%%%%%%%%%%%%%%%%%%%%%%%%%%%%%%%%%%%%%%%%%%%%%%%%%%%%%%%%%%%%%%
\begin{abstract}
%In visual terms, human-made environments typically contain low-textured scenes, where it is often difficult to find a sufficient number of point features. Due to this reason, the performance of visual odometry algorithms in these scenarios tends to degrade. As an alternative, other geometrical visual cues, such as lines, can be useful at this point. Additionally, these scenarios typically present structural regularities, such as parallelism or orthogonality, and hold the Manhattan World assumption~\cite{Coughlan1999Manhattan}. Under this context, in this work we introduce \approachname{}, a RGB-D visual odometry system that combines both point and line features and leverages, if exist, structural regularities and the Manhattan axes of the scene. Initially, these structural constraints are used to improve the accuracy of the 3D position of the extracted lines. Next, these constraints are combined with the estimated Manhattan axes and the reprojection errors of points and lines to improve the camera pose accuracy during a local map optimization stage. This combination enables our approach to operate even in the absence of these constraints, allowing to work in a wider variety of scenarios. Furthermore, we propose a novel multi-view Manhattan axes estimation procedure, which mainly rely on line features for this purpose. The proposed visual odometry system is validated using several popular datasets, outperforming other state-of-the-art solutions, and comparing favourably even with some SLAM approaches.
Visual odometry algorithms tend to degrade when facing low-textured scenes ---from e.g. human-made environments---, where it is often difficult to find a sufficient number of point features. Alternative geometrical visual cues, such as lines, which can often be found within these scenarios, can become particularly useful. Moreover, these scenarios typically present structural regularities, such as parallelism or orthogonality, and hold the Manhattan World assumption. Under these premises, in this work, we introduce \approachname{}, an RGB-D -based visual odometry approach that combines both point and line features and leverages, if exist, those structural regularities and the Manhattan axes of the scene. Within our approach, these structural constraints are initially used to estimate accurately the 3D position of the extracted lines. These constraints are also combined next with the estimated Manhattan axes and the reprojection errors of points and lines to refine the camera pose by means of local map optimization. Such a combination enables our approach to operate even in the absence of the aforementioned constraints, allowing the method to work for a wider variety of scenarios. Furthermore, we propose a novel multi-view Manhattan axes estimation procedure that mainly relies on line features. \approachname{} is assessed using several public datasets, outperforming other state-of-the-art solutions, and comparing favourably even with some SLAM methods.
\end{abstract}
% Keywords appear just beneath the abstract. Use only for final RAL version.
% \begin{IEEEkeywords}
% Mapping, Localization, Visual-Based Navigation
% \end{IEEEkeywords}
%%%%%%%%%%%%%%%%%%%%%%%%%%%%%%%%%%%%%%%%%%%%%%%%%%%%%%%%%%%%%%%%%%%%%%%%%%%%%%%%
\section{Introduction}

% Background 
%Visual Odometry (VO) is the process of estimating the trajectory of a camera within an environment by analysing the captured sequence of images. VO is a key part of a more sophisticated family of methods known as Simultaneous Localization and Mapping (SLAM), which typically combine VO with a loop closure detection approach to perform both tasks at the same time. When a previously seen place is revisited, the accumulated drift produced by VO can be minimized using an optimization procedure. However, this strategy does not completely remove the position error from the system and, therefore, the overall performance of any SLAM system is determined by the VO accuracy~\cite{Yang2017FeaturebasedOD}.
Visual Odometry (VO) is the process of estimating the trajectory of a camera within an environment by analysing the sequence of images captured. VO is a key part of a more sophisticated family of methods known as Visual Simultaneous Localization and Mapping (V-SLAM), which typically combine VO with a loop closure detection approach to perform both tasks at the same time. When a previously seen place is revisited, the accumulated drift produced by VO can be alleviated incorporating new constraints into the optimization stage. However, this strategy does not completely remove the camera pose error, so that the overall performance of any SLAM system gets determined by the VO accuracy~\cite{Yang2017FeaturebasedOD}.

Many VO and SLAM systems rely on point features because of their wider applicability in general terms~\cite{Mur2017ORBSLAM2}. However, in low-textured scenarios, their performance decrease due to the low number of points detected~\cite{Company2020lipo}. In this regard, the combination of point and line features has been demonstrated to reduce the number of tracking failures in these environments~\cite{Pumarola2017PLSLAM,gomezojeda2017plslam,Company2020lipo}. A complementary technique is to take profit of the structural constraints typically present in these scenarios, such as parallelism and/or orthogonality, through a pose-graph optimization strategy~\cite{Zhang2019PointPLane}. Another well-known strategy, which can be used to reduce the rotation drift in human-made environments, is to adopt the Manhattan World (MW) assumption~\cite{Coughlan1999Manhattan}. This hypothesis assumes a Cartesian coordinate system for the environment and that most part of the geometrical entities present in the scene align to one of its axes, named as Manhattan Axes (MA). This assumption is fundamentally used during the tracking stage~\cite{zhou2016divide, kim2017OPVO, kim2018LPVO, kim2018linear, Wang2019rgbdSLAM}. Nonetheless, these methods do not usually take into account that some indoor environments are not strictly adhering to this assumption, leading to degradation in accuracy or even to tracking failures~\cite{yunus2021manhattanslam}. Additionally, most of these works rely on planes to estimate and track the MA. These features lead to more accurate estimations of the MA than other features, such as lines, but they also mean higher computational times.

\begin{figure}[tb]
    \begin{center}
    \includegraphics[width=0.95\columnwidth, clip,trim={230 100 120 100}]{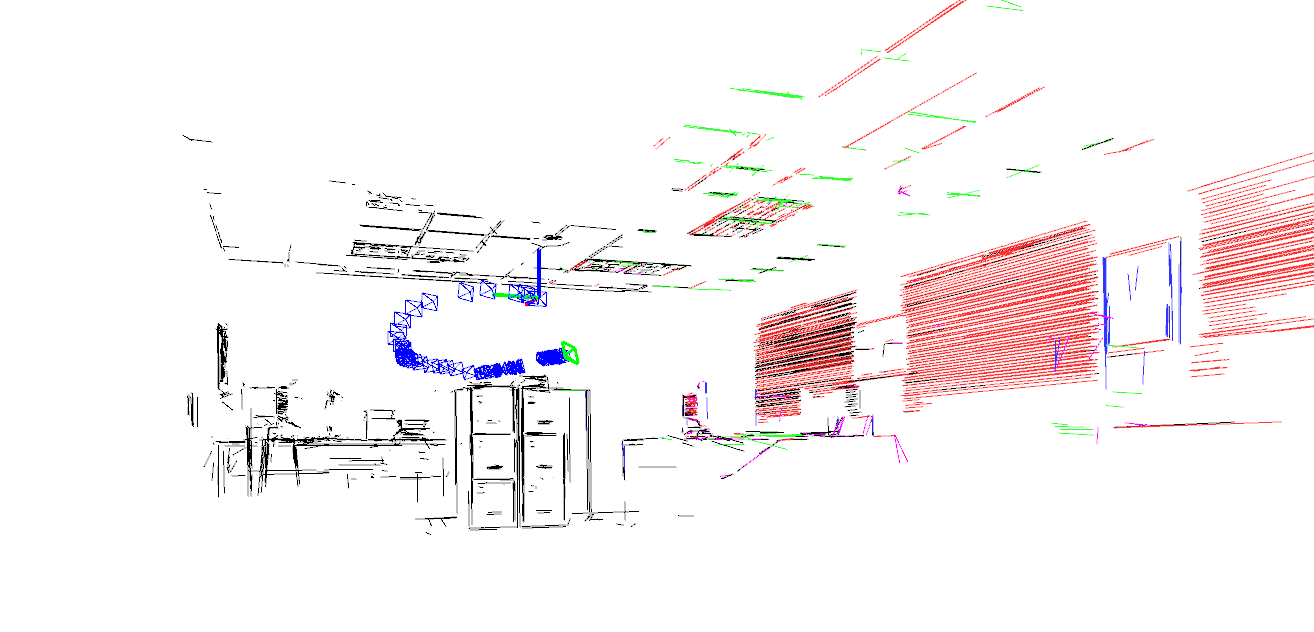}
    \end{center}
    %\caption{Overview of the generated local map using the proposed approach. For a better reader comprehension, only line features are displayed. As can be observed, human-made environment are typically rich in line features. Furthermore, parallel and orthogonal relations between lines are highly present due to the design of these environments. To conclude, a Manhattan axis line association is presented in red, green or blue, while non-associated lines are marked in purple. Whereas, those lines not included in the covisibility graph are marked in black. }
    \vspace{-0.4cm}
    \caption{Example of local map generated by \approachname{}. For a better understanding, only line features are shown. The map corresponds to a human-made environment, which, as expected, is rich in line features. Furthermore, parallel and orthogonal relations between lines are highly present due to the design of these environments. The Manhattan axis line associations are shown using red, green and blue colours, while non-associated lines are labelled in purple- Those lines not included in the covisibility graph are shown in black.}
    \label{fig:gen_overview}
    %\vspace{-0.65cm}
\end{figure}

% Overview of the proposed work
%On this basis, this work exploits the benefits of point and line features used in combination with structural constraints and a MA alignment to propose a new RGB-D VO framework called \approachnamelong{} (\approachname). As suggested, the proposed system relies on point and line features, due to their low extraction times. However, traditional depth extraction methods produce inaccuracies, which result from occlusions, depth discontinuities and RGB-D noise. These inaccuracies are even more evident in lines than in points. To alleviate this effect, we propose a two-step procedure, where, first, for each line detected in the image plane, its 3D line endpoints are estimated using a robust fitting procedure. Next, estimated endpoints are refined using the structural regularities that the scene presents. Additionally, our approach proposes a novel local map optimization stage which combines point and line reprojection errors along with structural regularities and a MA alignment, resulting into a more precise local trajectory estimation. Unlike other approaches, where the MW constraints are used during the tracking stage, our solution considers the MW assumption during this local map optimization, which allows us not to slow down the tracking, which typically requires operating in real-time. Finally, we propose a novel multi-view MA initialization procedure. An example of a result obtained using our approach is shown in Fig.\ref{fig:gen_overview}. In summary, the most important contributions of this work are:
Based on the above, this work exploits the benefits of point and line features used in combination with structural constraints and MA alignment to propose a new RGB-D VO framework named as \approachname{} from \emph{\approachnamelong{}}. As already said, the proposed method relies on point and line features, mostly because of their low detection times. Additionally, to address the inaccuracies in depth estimation which result from occlusions, depth discontinuities and RGB-D noise, which is even more notorious for lines than for points, we propose a two-step procedure that can be briefly stated as (1) for each line detected in the image plane, we estimate its 3D line endpoints using a robust fitting procedure, and (2) we next refine the estimated endpoints using the scene structural regularities. Moreover, our approach proposes a novel local map optimization stage which combines point and line reprojection errors along with structural regularities and MA alignment, resulting into more precise local trajectory estimations. Unlike other approaches, where the MW constraints are used during the tracking stage, our solution incorporates the MW assumption during local map optimization, which allows us not to slow down the tracking, which typically requires real-time operation to perform properly. Finally, we propose a novel multi-view MA initialization procedure. A first illustration of the performance of \approachname{} can be found in Fig.~\ref{fig:gen_overview}. 

In brief, the most important contributions of this work are:

% Contributions:
\begin{itemize}
    %\item A robust RGB-D VO framework for low-textured environments, which can improve the pose accuracy when structural regularities and MA alignment are present in the scene. Otherwise, our solution remains operational, as will be shown in the results.
    \item A robust RGB-D VO framework for low-textured environments, which can improve the pose accuracy when structural regularities and MA alignment are present in the scene. Otherwise, our solution remains operational, as will be shown in the experimental results section.
    %\item A 3D line endpoint extraction method based on the structural information present in the scene.
    \item A 3D line endpoint computation method based on the structural information present in the scene.
    \item  An accurate and efficient 3D local map optimization strategy, which combines reprojection errors with structural constraints and MA alignment.
    %\item A novel MA initialization procedure that refines the estimation of the traditional Mean Shift algorithm by using multiple frame observations in a multi-graph non-least square problem.
    \item A novel MA initialization procedure that refines the estimation of the traditionally employed Mean Shift algorithm by using multiple frame observations in a multi-graph non-linear least squares formulation.
    \item An extensive evaluation of the proposed approach on several public datasets and a comparison with other VO and SLAM state-of-the-art methods.
    %\item As an additional contribution, the source code of our approach is available online for the community\footnote{http://github.com/joanpepcompany/MSC-VO}.
    \item As an additional contribution, the source code \approachname{} is available online for the community\footnote{http://github.com/joanpepcompany/MSC-VO}.
    \end{itemize}
    
%The rest of the paper is organized as follows: Sec.~\ref{sec:relwork} overviews most important related works in the field; The proposed framework is introduced in Sec.~\ref{sec:overview}; Sec.~\ref{sec:results} reports on the results obtained; and, finally, Sec.~\ref{sec:conclusions} concludes the paper and suggests some future research lines.
The rest of the paper is organized as follows: Sec.~\ref{sec:relwork} overviews most relevant related works in the field; the proposed framework is introduced in Sec.~\ref{sec:overview}; Sec.~\ref{sec:results} reports on the results obtained; and, finally, Sec.~\ref{sec:conclusions} concludes the paper and suggests some future research lines.
    
%%%%%%%%%%%%%%%%%%%%%%%%%%%%%%%%%%%%%%%%%%%%%%%%%%%%%%%%%%%%%%%%%%
\section{Related Work}
\label{sec:relwork}
% Types of VO and SLAM methods and the problem of structured environments.
VO and Visual SLAM algorithms can be roughly classified into two main categories: feature-based and direct methods~\cite{Yang2017FeaturebasedOD}. Among them, feature-based approaches are typically more robust to illumination changes than direct methods. Despite their impressive results on well-textured scenarios~\cite{Mur2017ORBSLAM2}, their performance decreases when dealing with low-textured environments~\cite{Pumarola2017PLSLAM}. Due to this reason, some authors have opted to combine points with other geometric entities such as lines~\cite{Pumarola2017PLSLAM,gomezojeda2017plslam,Company2020lipo} or planes~\cite{Zhang2019PointPLane}.

Assuming a MW in human-made environments has demonstrated to be very effective to reduce the rotational drift ~\cite{zhou2016divide,kim2017OPVO,kim2018LPVO,kim2018linear,Wang2019rgbdSLAM}. Generally, this premise is taken into account during the tracking stage, being usually decoupled the rotation and the translation parts. Different strategies have been proposed to estimate and track the MA. For example, Zhou et al.~\cite{zhou2016divide} propose a single Mean Shift iteration that tracks the dominant MA for each frame by using a set of normal vectors. The translational part is computed through three simple 1D density alignments. In~\cite{kim2017OPVO}, the translation estimation is improved through a Kanade-Lucas-Tomasi (KLT) feature tracker. However, these two approaches require the existence of multiple orthogonal planes per frame. To solve this issue, Kim et al.~\cite{kim2018LPVO} combine line and plane features within a Mean Shift-based approach. In addition, they propose to use the reprojection error from the tracked points in the estimation of the translation. In a more recent work, they add an orthogonal plane detection and tracking method~\cite{kim2018linear}. Another solution to improve the tracking accuracy is presented in~\cite{Wang2019rgbdSLAM}, where the authors introduce the concept of plane orientation relevance to track the MA. More recently, other features are employed in~\cite{Li2021RGB-DSLAM}, which combines vanishing directions of 3D lines and plane normal vectors to track the MA. These works mainly rely on planar features or point normals, which, despite the accuracy that can be achieved, require more computational resources than other simpler features, such as lines. Moreover, the accuracy of the estimated MA determines the correctness of the system during its operation. To reduce these inaccuracies, Li et al.~\cite{li2019leveraging} describe a method that refines the reference MA by tracking it on each frame, and, thus, obtains multiple reference MA, which are later fused by Kalman Filtering. Following this idea, we propose to refine the position of 3D lines during MA estimation by using a graph-based non-linear error function that includes multiple views of the lines. This allows us to use them into an MA detection procedure.

% Back-end - Local Map optimization
%A local map optimization is usually performed in the back-end to reduce the errors produced during the tracking stage. In this regard, some approaches refine the pose of some previous frames after tracking the MA. For instance, in~\cite{Li2018AMonoSLAM}, the authors propose a line-based local optimization method to refine only the translation. However, the rotation is still computed using the decoupled tracking strategy. Moreover, other approaches~\cite{Zhang2019PointPLane, Li2021RGB-DSLAM} perform this local optimization by combining point and plane features in conjunction with structural constraints, which have been shown to achieve better results than the decoupled scheme~\cite{Li2021RGB-DSLAM}.
%% It might be said that the combination of multiple observations with structural constraints improves notably the inaccuracies of the decoupled tracking strategy. 
Local map optimization is usually performed in the back-end to reduce the errors produced during the tracking stage. In this regard, some approaches refine the pose of some previous frames after tracking the MA. For instance, in~\cite{Li2018AMonoSLAM}, the authors propose a line-based local optimization method to refine only the translation. However, the rotation is still computed using the decoupled tracking strategy. Moreover, other approaches~\cite{Zhang2019PointPLane, Li2021RGB-DSLAM} perform this local optimization by combining point and plane features in conjunction with structural constraints, which have been shown to achieve better results than the decoupled scheme~\cite{Li2021RGB-DSLAM}.

% Ausencia de manhattan axes.
%There exist indoor environments that do not strictly conform to the MW assumption. In these cases, the performance of approaches purely based on it decreases, producing even tracking failures. To overcome this issue, Zhang et al.~\cite{Zhang2019PointPLane} propose using parallel and perpendicular constraints as an alternative to the MW assumption. Despite its advantages, this method can not reduce the long-term rotation error as the MW assumption does. Another solution is presented in~\cite{yunus2021manhattanslam}, where the authors use either a decoupled or non-decouple tracking strategy depending on whether the MW assumption is followed in the scene. These strategies provide to these works to not only focus on a specific environment.
There exist indoor environments that do not strictly conform to the MW assumption. In these cases, the performance of approaches purely based on it degrades, even leading to tracking failures. To overcome this issue, Zhang et al.~\cite{Zhang2019PointPLane} propose using parallel and perpendicular constraints as an alternative to the MW assumption. Despite its advantages, this method can not reduce the long-term rotation error as the MW assumption does. Another solution is proposed in~\cite{yunus2021manhattanslam}, where the authors use either a decoupled or a non-decoupled tracking strategy depending on whether the scene meets the MW assumption. These strategies permit these works to not only focus on a specific environment.

% Resumen de la sección
%To sum up, as the reviewed related work suggest, the use of the MW assumption improves the localization accuracy of VO and SLAM methods. However, using this assumption, as a principal source of the tracking procedure, causes a failure in some scenes where the MW assumption does not adhere, and therefore, these solutions are addressed to a specific environment. As a solution, we propose the use of the MA in a local map optimization, allowing its absence and a further MA localization. Additionally, in ~\cite{Li2021RGB-DSLAM}, the authors demonstrate that this assumption increase the performance of their method. However, as the results suggest, the most noticeable improvement is obtained when using the structural constraints in the local map optimization. To this end, we propose a novel local map optimization that combines: the point and line reprojection error, the MA alignment and the structural constraints of the scene. Allowing that, the absence of some of these constraints in the scene do not affect to the overall performance. As a result, the proposed work increase the localization accuracy, regarding the reviewed VO, and it allows working in a wider range of scenarios.   
The related works reviewed above suggest the use of the MW assumption to increase the localization accuracy of VO and SLAM methods. However, using this assumption as a primary source in the tracking procedure can lead to failures in some scenes where the MW assumption is not satisfied, what can restrict those solutions for certain specific environments. As a solution, we propose the incorporation of the MA in local map optimizations. Additionally, we take inspiration from~\cite{Li2021RGB-DSLAM}, which reports the structural constraints as beneficial for the pose refinement process. To this end, we propose a novel local map optimization approach that combines the point and line reprojection error, the MA alignment and the structural constraints of the scene. Allowing that, the punctual dissatisfaction of some of these constraints does not affect the overall performance. As a result, our method leads to higher localization accuracy and allows working in a wider range of scenarios.   

%%%%%%%%%%%%%%%%%%%%%%%%%%%%%%%%%%%%%%%%%%%%%%%%%%%%%%%%%%%%%%%%%%%%%%%%%%%%%%%%%%%%%
\section{\approachname{} Overview}
\label{sec:overview}

\begin{figure}[tb]
    \begin{center}
    \includegraphics[width=0.98\columnwidth, clip,trim={0 0 0 0}]{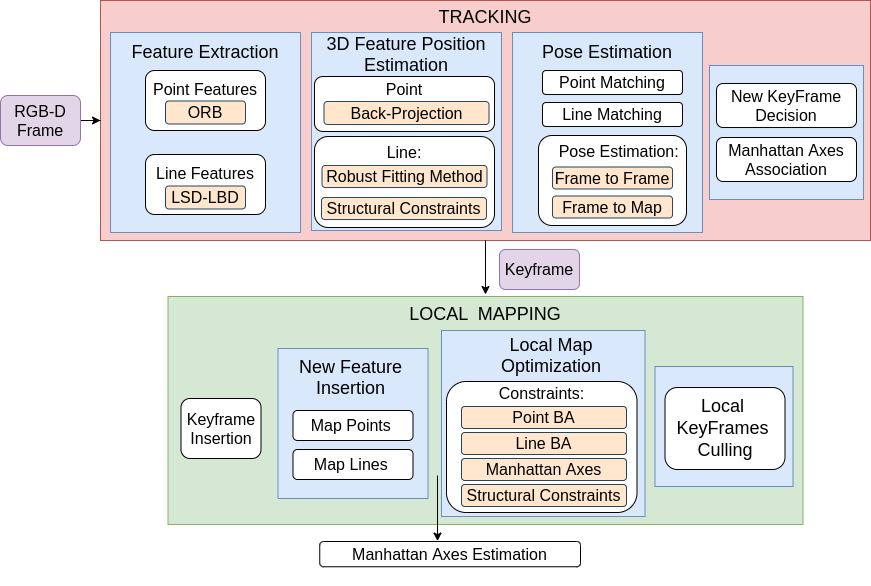}
    \end{center}
    \vspace{-4mm}
    %\caption{\approachname{} overview}
    \caption{Overview of \approachname{}}
    % \vspace{-0.65cm}
    \label{fig-pipeline}
\end{figure}

%The proposed approach is built on top of the tracking and local mapping components of ORB-SLAM2~\cite{Mur2017ORBSLAM2}. Consequently, the system consists of two threads running in parallel, as it is illustrated in Fig.~\ref{fig-pipeline}. Further details of each of these modules can be found next.
\approachname{} is built on top of the tracking and local mapping components of ORB-SLAM2~\cite{Mur2017ORBSLAM2}. Therefore, it comprises two threads running in parallel, as it is illustrated in Fig.~\ref{fig-pipeline}. Further details on \approachname{} can be found next.

\subsection{Tracking}
%The tracking thread is in charge of estimating the position of every captured frame. Additionally, this module decides whether a new keyframe needs to be created and associate, if possible, each new map line with one of the MA.
The tracking thread is in charge of estimating the position of every frame captured. Additionally, this module decides whether a new keyframe needs to be created. It also associates each new map line with one of the MA, if possible.

\subsubsection{Feature Extraction}
%Every received frame $I_t$ coming from the RGB-D sensor at time $t$ consists of a colour $I^c_t$ and a depth image $I^d_t$. Then, point and line features are extracted from $I^c_t$. Points are detected and described using ORB~\cite{rublee2011orb}, while lines are detected using Line Segment Detector (LSD)~\cite{Grompone2010LSD} and described using the binary form of the Line Band Descriptor (LBD)~\cite{ZHANG2013LBD}. The location of a point $i$ in image coordinates is denoted as $p_i$. Conversely, each line segment $j$ detected in the image plane is represented by a start point $s_j$ and an end point $e_j$. Additionally, the normalized line equation $l_j$ can be computed as:
Every frame $I_t$ coming from the RGB-D sensor at time $t$ consists of a colour image $I^c_t$ and a depth image $I^d_t$. Point and line features are extracted from $I^c_t$. Points are detected and described using ORB~\cite{rublee2011orb}, while lines are detected using the Line Segment Detector (LSD)~\cite{Grompone2010LSD} and described using the binary form of the Line Band Descriptor (LBD)~\cite{ZHANG2013LBD}. In the following, the location of a point $i$ in image coordinates is denoted as $p_i$, while each line segment $j$ detected in the image plane is represented by a start point $s_j$ and an end point $e_j$. Additionally, the normalized line $l_j$ is expressed as:
\begin{equation}
l_j=\frac{e_j - s_j}{\left\| e_j - s_j\right\|}\,.
\label{eq:line-coef}
\end{equation}

\subsubsection{3D Feature Position Estimation}
%Once points and lines have been detected and described, their 3D positions in camera coordinates are obtained. A point $p_i$ is backprojected using as depth the correspondent value of its 2D position in $I^d_t$. The resulting 3D position is denoted as $P_i^c$. Conversely, lines are more affected than points by in-depth discontinuities and occlusions, and, therefore, this simple procedure can end up with inaccurate 3D lines. To reduce this effect, we propose a robust two-step method to compute the 3D line end points.
Once points and lines have been detected and described, their 3D positions in camera coordinates are obtained. A point $p_i$ is backprojected using as depth the value corresponding to its 2D position in $I^d_t$. The resulting 3D position in camera coordinates is denoted as $P_i^c$. Since lines are more affected than points by depth discontinuities and occlusions, this simple procedure can end up with inaccurate 3D lines. To reduce this effect, we propose a robust two-step method to compute the 3D line endpoints.

%First, for every line $j$, we calculate an initial 3D position of its end points, denoted by $\{S_j^c,E_j^c\}$, by backprojecting a subset of the points that conforms the line in the image and, next, performing a robust fitting method as in~\cite{Li2021RGB-DSLAM}. The 3D normalized line equation $L_j^c$ are computed similarly to Eq.~\ref{eq:line-coef}. Next, structural constraints of the scene are employed to refine each detected line. We start by associating parallel and perpendicular lines. To this end, for every possible pair of lines $(L_m^c, L_n^c)$ detected in the current image, we compute the cosine of the angle formed by the two vectors as:
First, for every line segment $j$, we calculate an initial 3D position for its endpoints, denoted by $\{S_j^c,E_j^c\}$, by backprojecting a subset of the points that conforms the line in the image and, next, performing a robust fitting step as in~\cite{Li2021RGB-DSLAM}. The 3D normalized line $L_j^c$ is computed similarly to Eq.~\ref{eq:line-coef}. Next, we employ the structural constraints of the scene to refine each detected line. We start by associating parallel and perpendicular lines. To this end, for every possible pair of lines $(L_m^c, L_n^c)$ detected in the current image, we compute the cosine of the angle between the two direction vectors by means of the dot product:
\begin{equation}
\cos \left(L_m^c,L_n^c\right) = \frac{L_m^c \cdot L_n^c}{\|L_m^c\|\|{L_n^c}\|}\,.
\label{eq:angle-btwn-lines}
\end{equation}

%We choose only those pairs $(L_m^c, L_n^c)$ whose cosine value is close to 0 or 1 representing, respectively, perpendicular or parallel lines. The selected pairs are employed to refine their line end points by means of a non-linear graph-based optimization procedure. We use the Levenberg–Marquardt algorithm implemented in g2o~\cite{Kummerle2011g2o} to this end.  Formally, we define the angle difference $d$ between two lines $L_m^c$ and $L_n^c$ as:
We choose only those pairs $(L_m^c, L_n^c)$ whose cosine value is close to 0 or 1 representing, respectively, perpendicular or parallel lines. The selected pairs are employed to refine their line endpoints by means of non-linear optimization. We use the Levenberg–Marquardt algorithm implemented in g2o~\cite{Kummerle2011g2o} to this end. Formally, we define the orientation discrepancy $d$ between lines $L_m^c$ and $L_n^c$ as:
\begin{equation}
%    d(L_m^c, L_n^c) = \frac{L_m^c \cdot L_n^c}{\|L_m^c\|}\,.
    d(L_m^c, L_n^c) = |\cos \left(L_m^c,L_n^c\right)|\,.
\label{eq:angdiff}
\end{equation}

%Let us denote $\mathbb{L_{\perp}}$ and $\mathbb{L_{\parallel}}$ as the sets of valid perpendicular and parallel line pairs, respectively. Given a pair $(L_m^c, L_n^c) \in \mathbb{L_{\perp}}$, the error term $\mathbf{E}^{\perp}_{m, n}$ is defined as:
Let us denote $\mathbb{L_{\perp}}$ and $\mathbb{L_{\parallel}}$ as the sets of, respectively, valid perpendicular and valid parallel line pairs. Given a pair $(L_m^c, L_n^c) \in \mathbb{L_{\perp}}$, the error term $\mathbf{E}^{\perp}_{m, n}$ is defined as:
\begin{equation}
\label{eq-perp_error}
%    \mathbf{E}^{\perp}_{m, n} = \left\|\,d(L_m^c, L_n^c)\,\right\|_{\Sigma}^{2}\,,
    \mathbf{E}^{\perp}_{m, n} = d(L_m^c, L_n^c) \cdot \omega_n^{-1},
\end{equation}
%where $\Sigma$ is the covariance matrix of the line response returned by the LSD algorithm. Similarly, for another pair $(L_m^c, L_n^c) \in \mathbb{L_{\parallel}}$, the error term $\mathbf{E}^{\parallel}_{m, n}$ is defined as follows:
where $\omega_n$ weights the error term in accordance to the line response returned by the LSD algorithm for segment $n$. Similarly, for another pair $(L_m^c, L_n^c) \in \mathbb{L_{\parallel}}$, the error term $\mathbf{E}^{\parallel}_{m, n}$ is defined as follows:
\begin{equation}
\label{eq-par_error}
%    \mathbf{E}^{\parallel}_{m, n} = \left\|\,1 - d(L_m^c, L_n^c)\,\right\|_{\Sigma}^{2}\,.
%    \mathbf{E}^{\parallel}_{m, n} = \left(1 - d(L_m^c, L_n^c)\,\right) \cdot \omega_n^{-1}\,,
    \mathbf{E}^{\parallel}_{m, n} = \sqrt{1 - d^{\,2}(L_m^c, L_n^c)} \cdot \omega_n^{-1}\,.
\end{equation}
where $d(\cdot,\cdot) \in [0,1]$. 

%\begin{equation}
%\label{eq-par_error}
%    \mathbf{E}^{\parallel}_{m, n} = \left(1 - d(L_m^c, L_n^c)\,\right)^T \cdot \omega_n^{-1}\,.
%\end{equation}

\begin{figure}[tb]
    \centering
    \begin{tabular}{@{\hspace{0mm}}c@{\hspace{5mm}}c@{\hspace{0mm}}}
    \includegraphics[width=0.47\linewidth, clip=true, trim=0 0 10 0]{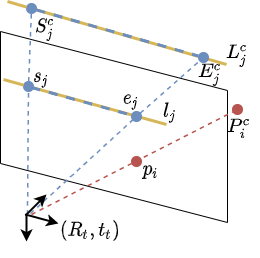} & 
    \includegraphics[width=0.4\linewidth, clip=true, trim= 0 0 5 0]{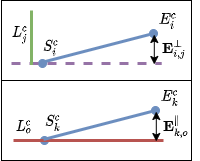} \\
    \end{tabular}
%  \vspace{-4.5mm}
    %\caption{Image left illustrates the 2D and 3D features notations in a frame. Image right represent the line endpoints error terms. $S^c$ and $Q^c$ represent the line endpoints to optimize. The right-top image illustrates the error term $E_{c,f}$ which is obtained by the angle between the normalized line equation obtained by the line endpoints and the perpendicular line, shown in purple, of a perpendicular association $L^f$.  The right-bottom image illustrates the parallel error term, where the error term is obtained between the normalized line coefficient, and its parallel association  $L^e$.}
    \caption{The left drawing illustrates the notation used for 2D and 3D features, while the right drawings illustrate the line endpoints error terms. $S^c_j$ and $E^c_j$ are the line endpoints to optimize. The right-top drawing shows the error term $\mathbf{E}_{i,j}^{\perp}$ as the cosine of the angle between the normalized line defined by $S^c_i$ and $E^c_i$ and a perpendicular line, shown in green, for a perpendicular association $L^c_j$. The right-bottom drawing illustrates the parallel error term $\mathbf{E}_{k,o}^{\parallel}$, calculated as the sine of the angle between the normalized line defined by $S^c_k$ and $E^c_k$ and a parallel association $L^c_o$. (Both cases assume $\omega = 1$.)}
    % \vspace{-0.65cm}
    \label{fig:notation}
\end{figure}

We define $\mathbf{L}$ as the set of variables to be optimized, which includes those lines that have at least one structural association either on $\mathbb{L_{\perp}}$ or $\mathbb{L_{\parallel}}$. We then compute the optimal line end points of $\mathbf{L}$ by minimizing the following cost function:
\begin{equation}
    \mathbf{L} = \underset{\mathbf{L}}{\operatorname{argmin}} \left( \sum_{(i,j) \in \mathbb{L_{\perp}}} \rho\left( \mathbf{E}^{\perp}_{i, j} \right) + \sum_{(k,o) \in \mathbb{L_{\parallel}}} \rho\left( \mathbf{E}^{\parallel}_{k, o} \right) \right)\,,
\end{equation}
%being $\rho$ a Huber loss function  to  reduce  the  influence  of outliers. Figure \ref{fig:notation} summarizes: the notation of points and lines regarding frame coordinates, and the two error terms defined in this section. To sum up, as it will be shown in Sec.~\ref{sec:results}, using this procedure, the line estimation accuracy is improved, benefiting the whole system. 
where $\rho$ is the Huber loss function to reduce the influence of outliers. Figure~\ref{fig:notation} summarizes the notation of points and lines regarding frame coordinates, and the two error terms defined in this section. As it will be shown in Sec.~\ref{sec:results}, using the outlined procedure, the 3D lines estimation accuracy improves, benefiting the whole system. 

% The error term of a line with a parallel association is given by:  
% \begin{equation}
% \label{eq-par_error}
%     \mathbf{E}_{c, e} = \left\|1 - \left(\frac{(Q^c - S^c) \cdot L^{e} }{\left\| Q^c - S^c\right\| }\right )\right\|_{\Sigma}^{2},
% \end{equation}

% where $\Sigma$ is the covariance matrix.

% Whereas, the error term of a line with its perpendicular association is:
% \begin{equation}
% \label{eq-perp_error}
%     \mathbf{E}_{c, f} = \left\| \left(\frac{(Q^{c} - S^{c}) \cdot L^{f} }{\left\| Q^{c} - S^{c}\right\|}\right )\right\|_{\Sigma}^{2}.
% \end{equation}

\subsubsection{Pose Estimation}
Once features are extracted, an optimization procedure is carried out to estimate the current camera orientation $\mathbf{R}_t \in SO(3)$ and translation $\mathbf{t}_t \in \mathbb{R}^3$. Initially, map points and lines observed in the previous frame are projected to the current frame, assuming a constant velocity motion model. Next, two sets of 2D-3D correspondences, one for points as in~\cite{Mur2017ORBSLAM2} and one for lines as in~\cite{gomezojeda2017plslam}, are computed. These associations are then employed to optimize the current camera pose, minimizing the following cost function:
\begin{equation}
    \{\mathbf{R_t, t_t}\} = \underset{\mathbf{R_t, t_t}}{\operatorname{argmin}} \left( \sum_{i \in \mathbb{P}} \rho\left( \mathbf{E}^{p}_{i} \right) + \sum_{j \in \mathbb{V}} \rho\left( \mathbf{E}^{l}_{j} \right) \right)\,,    
    \label{eq:poseest}
\end{equation}
%where $i \in \mathbb{P}$ and $j \in \mathbb{V}$ are, respectively, the sets of all point and line matches. The error term for the observation of a map point $i$ is defined as:
where $\mathbb{P}$ and $\mathbb{V}$ are, respectively, the sets of all point and line matches. The error term for the observation of a map point $i$ is defined as:

\begin{equation}
%    \mathbf{E}^{p}_{i} = \left\|\,p_i - \pi(R_t P_i^w + t_t)\,\right\|_{\Sigma}^{2}\,,
    \mathbf{E}^{p}_{i} = \left\|\,p_i - \pi(R_t P_i^w + t_t)\,\right\|^{2} \cdot \omega_i^{-1}\,,
    \label{eq:errorreprpoint}
\end{equation}
%being $P_i^w \in \mathbb{R}^3$ the corresponding point in world coordinates of $p_i \in \mathbb{R}^2$. The projection function $\pi$ transforms a 3D point $P_i^c$ in camera coordinates into the image plane using camera calibration~\cite{Hartley2003}. Conversely, the error term for an observed map line $j$ in the current frame is defined as:
where $P_i^w \in \mathbb{R}^3$ is the point in world coordinates corresponding to $p_i \in \mathbb{R}^2$ and $\omega_i$ weights the error term in accordance to the response of the ORB detector. The projection function $\pi$ transforms a 3D point $P_i^c$ in camera coordinates into the image plane using the camera calibration parameters~\cite{Hartley2003}. On the other side, the error term for an observed map line $j$ in the current frame is defined as:
%\begin{equation}
%    \mathbf{E}^{l}_{j} = \left\|\,l_j \cdot h(\pi(R_t S_j^w + t_t)) + l_j \cdot h(\pi(R_t E_j^w + t_t))\,\right\|_{\Sigma}^{2}\,,
%    \label{eq:errorreprline}
%\end{equation}
%\begin{equation}
%    \mathbf{E}^{l}_{j} = \left(\left(n_j \cdot \pi(R_t S_j^w + t_t)\right)^2 + \left(n_j \cdot \pi(R_t E_j^w + t_t)\right)^2\right) \cdot \omega_j^{-1}\,,
%    \label{eq:errorreprline2}
%\end{equation}
\begin{equation}
    \mathbf{E}^{l}_{j} = \left\|n_j \cdot \pi(R_t S_j^w + t_t),\ n_j \cdot \pi(R_t E_j^w + t_t)\right\|^2 \cdot \omega_j^{-1}\,,
    \label{eq:errorreprline}
\end{equation}
%being $L_j^w = \{S_j^w, E_j^w\}$ the map line, in world coordinates, that matches the 2D segment $l_j$ and $h$ is a function that transforms a point to homogeneous coordinates. Once the camera pose is estimated, we project the local map into the current frame to obtain more correspondences, as performed in~\cite{Mur2017ORBSLAM2}. The pose is optimized again with the resulting matches.
where $L_j^w = \{S_j^w, E_j^w\}$ is the map line in world coordinates that matches the 2D segment $l_j$ with normal vector $n_j$. Once the camera pose has been estimated, we project the local map into the current frame to obtain more correspondences, as performed in~\cite{Mur2017ORBSLAM2}. The pose is optimized again with the resulting matches.

\subsubsection{Keyframe Insertion}
%When the camera pose is estimated, the current frame is evaluated to decide if it should be considered as a new keyframe. We use a similar policy as ORB-SLAM2~\cite{Mur2017ORBSLAM2}, but considering line correspondences in the procedure. Unlike ORB-SLAM2, we do not use the condition of a minimum features tracked. The rationale behind this idea is that the proposed method is focused on low-textured environments, where typically the number of features tracked per frame drastically changes between scenes. Therefore, it is not possible to fix a reasonable threshold. In its place, we propose to use the ratio between the correct tracked features and the total number of features. Once a new keyframe is generated, points and lines are included in the local map and redundant features are culled, as performed in~\cite{Mur2017ORBSLAM2}. For each new map line, we search for parallel or perpendicular line correspondences in the local map. Additionally, each line is then associated to an MA, if possible, as explained in the next section.
Once the camera pose has been estimated, the current frame is evaluated to decide whether it should be considered as a new keyframe. We use a similar policy as ORB-SLAM2~\cite{Mur2017ORBSLAM2}, but incorporating line correspondences. Unlike ORB-SLAM2, we do not use the condition of a minimum number of features tracked. The rationale behind this idea is that the proposed method is focused on low-textured environments, where typically the number of features tracked per frame can change drastically between scenes. Therefore, it is not possible to fix a reasonable threshold. Instead, we propose to use the ratio between the current frame features that are being tracked in the map, and the sum of these features with the ones that could be potentially created. Once a new keyframe is generated, points and lines are included in the local map and redundant features are culled, as performed in~\cite{Mur2017ORBSLAM2}. For each new map line, we search for parallel or perpendicular line correspondences in the local map. Additionally, each line is also associated to an MA, if possible, as explained in the next section.

\subsubsection{Manhattan Axes Association}
\label{subsection-ManhAxesAssoc}
%When a keyframe is inserted, new map lines are associated to one of the MA, if available. To this end, we compare every line $L_j^w$ with each of the three axis. If the cosine of the angle between them, computed using Eq.~\ref{eq:angle-btwn-lines}, is close to $1$, the line can be considered as parallel to the axis, and then they are matched. These associations will be used during the local map optimization to reduce the camera rotation drift. Notice that, given the combination of structural constraints and this MA alignment, our approach is able to operate even if these axes are not available. The procedure to estimate these MA is explained in Sec.~\ref{section-manh_axes}.
Given $\mathcal{M} = \{\text{MA}_0,$ $\text{MA}_1, \text{MA}_2\}$ as the set of Manhattan Axes, when a new keyframe is inserted, each new map line $j$ is associated to axis $M_j \in \mathcal{M}$ whenever possible. To this end, we compare every line $L_j^w$ with each of the three axes: if the value of expression in Eq.~\ref{eq:angdiff} gets close enough to 1 for axis MA$_k$, the line is considered as parallel to MA$_k$, and they are matched, i.e. $M_j = \text{MA}_k$. These associations are used during local map optimization to reduce the camera rotation drift. Notice that, given the combination of structural constraints and this MA alignment, our approach is able to operate even if these axes are not available. The procedure to estimate these MA is explained in Sec.~\ref{section-manh_axes}.

%After a keyframe insertion, we evaluate the correspondences between the created map lines and one of the MA. As stated previously, we only use the MW assumption in the local map optimization because of the advantages pointed out above. However, it may be the case that MA reference is not extracted yet or do not appear in the scene. For this reason, some keyframes do not benefit of this information, but, as an advantage, it does not affect to the tracking procedure. Due to the MA estimation is processed in the local mapping thread, this is further explained in Section \ref{section-manh_axes}. The following describes the procedure to evaluate the correspondence between a line to a single Manhattan axis. To this end, we compare every created map line with the three MA. Then, a correspondence is found, when the $cos$ of the angle between these two vectors, which is computed using the equation~\ref{eq:angle-btwn-lines}, is around $1$. Meaning that these two vectors are almost parallel. Once, these associations are extracted, the local map optimization use them to reduce the long-term camera rotation drift. 

\subsection{Local Mapping}
% Whenever a keyframe $K_i$ is inserted, the local mapping thread refines recent keyframe poses and landmarks by a multi-graph optimization process. Furthermore, this thread also estimates the reference MA, if required. Further details can be found next.
Whenever a keyframe is inserted, the local mapping thread refines recent keyframe poses and landmarks by a multi-graph optimization process. Furthermore, this thread also estimates the reference MA, if required. Finally, redundant keyframes are culled using the strategy introduced in~\cite{Mur2017ORBSLAM2}. Further details can be found next.

\subsubsection{Local Map Optimization}
\label{section-localmap}
%Once a keyframe $K_i$ is generated, the local optimization procedure refines its pose along with the poses of a set of connected keyframes $\mathcal{K}_c$ obtained from a covisibility graph~\cite{Mur2017ORBSLAM2} and all the map points $\mathcal{P}$ and lines $\mathcal{L}$ seen by those keyframes. Additionally, keyframes that observe these points and lines but are not connected to $K_i$, denoted by $\mathcal{K}_f$, are included in the optimization, but their poses remain fixed. We denote $\mathbb{P}_k$ and $\mathbb{V}_k$ as the sets of matches between, respectively, points and lines in $\mathcal{P}$ and $\mathcal{L}$ and features in a keyframe $k$. To introduce the structural constraints of the scene into our optimization problem, let us define $\mathbb{L}_{\perp}^k$ and $\mathbb{L}_{\parallel}^k$ as the sets of perpendicular and parallel pairs of lines in $\mathcal{L}$, respectively, co-observed in a keyframe $k$. Finally, we denote the set of map lines associated to a MA as $\mathbb{M}$. Defining $\Gamma = \{P_i^w, L_j^w, R_l, t_l,|i \in \mathcal{P}, j \in \mathcal{L}, l \in \mathcal{K_c}\}$ as the set of variables to be estimated, the optimization problem is defined as:
Once keyframe $k$ is generated, the local optimization procedure refines its pose along with the poses of a set of connected keyframes $\mathcal{K}_c$ obtained from a covisibility graph~\cite{Mur2017ORBSLAM2} and all the map points $\mathcal{P}$ and lines $\mathcal{L}$ seen by those keyframes. Other keyframes that observe these points and lines but are not connected to $k$, denoted by $\mathcal{K}_f$, are included in the optimization, but their poses remain fixed. We denote $\mathbb{P}_k$ and $\mathbb{V}_k$ as the sets of matches between, respectively, points and lines in $\mathcal{P}$ and $\mathcal{L}$ and features in keyframe $k$. To introduce the structural constraints of the scene into the optimization, we define $\mathbb{L}_{\perp}^k$ and $\mathbb{L}_{\parallel}^k$ as the sets of perpendicular and parallel pairs of lines in $\mathcal{L}$, respectively, co-observed in keyframe $k$. Finally, we denote as $\mathbb{M}$ the set of map lines that are associated to a MA and that are seen by any keyframe in $\mathcal{K}_c$. Defining $\Gamma = \{P_i^w, L_j^w, R_l, t_l,|i \in \mathcal{P}, j \in \mathcal{L}, l \in \mathcal{K}_c\}$ as the set of variables to be estimated, the optimization problem is defined as:
% \begin{equation}
% \begin{split}
%      \mathbf{\Gamma} = \underset{\mathbf{\Gamma}}{\operatorname{argmin}}
%     %\left[
%          \sum_{k \in \{\mathcal{K}_c \cup \mathcal{K}_f\}} \left( \sum_{x \in \mathbb{P}_k} \rho\left( \mathbf{E}^{p}_{x} \right) + \sum_{y \in \mathbb{V}_k} \rho\left( \mathbf{E}^{l}_{y} \right)\right) & \\
%          + \sum_{z \in \mathcal{K}_c} \left( \sum_{(i,j) \in \mathbb{L}_{\perp}^z} \rho\left( \mathbf{E}^{\perp}_{i, j} \right) + \sum_{(o,q) \in \mathbb{L}_{\parallel}^z} \rho\left(\mathbf{E}^{\parallel}_{o,q} \right)\right) & \\
%          + \sum_{w \in \mathbb{M}} \rho\left(\mathbf{E}_{w}^{M} \right)
%     %\right]
% \end{split}
% \end{equation}
\begin{equation}
\begin{split}
     \mathbf{\Gamma} = \underset{\mathbf{\Gamma}}{\operatorname{argmin}}
    \left[
         \sum_{k \in \{\mathcal{K}_c \cup \mathcal{K}_f\}} \left( \sum_{i \in \mathbb{P}_k} \rho\left( \mathbf{E}^{p}_{i} \right) + \sum_{j \in \mathbb{V}_k} \rho\left( \mathbf{E}^{l}_{j} \right)\right) 
    \right. & \\
         + \sum_{z \in \mathcal{K}_c} \left( \sum_{(i,j) \in \mathbb{L}_{\perp}^z} \rho\left( \mathbf{E}^{\perp}_{i, j} \right) + \sum_{(i,j) \in \mathbb{L}_{\parallel}^z} \rho\left(\mathbf{E}^{\parallel}_{i,j} \right)\right) 
         & \\
    \left.
         + \sum_{j \in \mathbb{M}} \rho\left(\mathbf{E}_{j,M_j}^{\parallel} \right)
    \right]
\end{split}
\end{equation}
%where $\mathbf{E}^{p}_{i}$, $\mathbf{E}^{l}_{j}$, $\mathbf{E}^{\perp}_{m, n}$ and $\mathbf{E}^{\parallel}_{m, n}$ were respectively defined in Eqs.~\ref{eq:errorreprpoint}, ~\ref{eq:errorreprline}, ~\ref{eq-perp_error} and~\ref{eq-par_error} and the MA alignment error $\mathbf{E}_{w}^{M}$ is the angle between a map line and its corresponding Manhattan axis, computed using Eq.~\ref{eq:angdiff}.
where $\mathbf{E}^{\perp}_{i, j}$, $\mathbf{E}^{\parallel}_{i, j}$, $\mathbf{E}^{p}_{i}$ and $\mathbf{E}^{l}_{j}$ were respectively defined in Eqs.~\ref{eq-perp_error},~\ref{eq-par_error},~\ref{eq:errorreprpoint} and~\ref{eq:errorreprline},  and the MA alignment error $\mathbf{E}_{j,M_j}^{\parallel}$ is the error term corresponding to a map line $j$ and its associated Manhattan axis $M_j \in \mathcal{M}$, calculated using Eq.~\ref{eq-par_error}.

\subsubsection{Manhattan Axes Estimation}
\label{section-manh_axes}
As already said, the Manhattan Axes comprise a set of three orthogonal directions, in world coordinates, which represent the main scene directions. These directions remain fixed over time and, therefore, their estimation should be very accurate to prevent misalignments during optimization steps. In this respect, this work proposes a coarse-to-fine MA estimation strategy, where the estimation at the coarsest level is obtained extending the work by Kim et al.~\cite{kim2018LPVO}. The estimated MA are then refined by considering multiple line observations along different keyframes.

% the coarse MA extraction is computed using the Mean Shift algorithm, as detailed in~\cite{kim2018LPVO}. However, unlike the original work, we use the 3D normalized line direction and the point normals of the scene. Regarding point normals, these are obtained using integral images as explained in~\cite{Holz2011Realtime}. In more detail, Holz et al. use integral images as a previous step to detect planes in the image. However, to reduce the computational time, we discretize the image by only selecting one point every ten rows and ten columns, which allows to know the main directions of the scene. Note that, the computation of point normals is only performed until a valid coarse MA is estimated. However, this coarse strategy is affected by the point normals and the line depth noise from a single frame. Thisffect is reduced by a multi-view optimization strategy.
% In a first stage, when the first keyframe $K_0$ is received, an initial MA are computed from it using the Mean Shift-based method proposed in~\cite{kim2018LPVO}, but just using the line director vectors and point normals of the scene as features. These point normals are calculated using a modified approach of the one proposed in~\cite{Holz2011Realtime}, which is based on integral images for a rapid normal detection. However, we only select points in a grid over the image for an even faster computation. This procedure is performed iteratively for each received keyframe until a valid, but typically noisy, MA are estimated. These MA will be refined by the multi-view optimization detailed next.
For a start, a first estimation of the MA is computed from the first keyframe once it is available using the Mean Shift-based method proposed in~\cite{kim2018LPVO}. In this first stage, the only features involved are the line direction vectors and the surface normal vectors for a selection of points defined over a grid. The normal vectors are calculated using a modified version of the approach proposed in~\cite{Holz2011Realtime}, which is based on integral images to speed up calculations. This procedure is repeated for the next keyframes until valid, though typically noisy, MA are obtained. 

Once the local map comprises a sufficient number of keyframes, being denoted by $\mathcal{K}_M$, a non-linear optimization procedure is performed in a second MA refinement stage, using hence the inaccurate MA computed in the first stage as initial guess. Given $\mathcal{M}$ as the set of MA, and defining $\mathbb{V}_k^{\text{MA}_i}$ as the set of map lines associated to the Manhattan axis MA$_i$ observed in keyframe $k$, the optimization problem can be stated as follows:
% \begin{equation}
% \begin{split}
%     \{\mathcal{M}\} = \underset{\mathcal{M}}{\operatorname{argmin}} 
%         \sum_{k \in \mathcal{K}_M}
%         \left(
%             \sum_{i \in \mathbb{V}_k^{M_0}} \rho(\mathbf{E}_{i}^{M_0})
%           + \sum_{j \in \mathbb{V}_k^{M_1}} \rho(\mathbf{E}_{j}^{M_1}) & \\
%           + \sum_{w \in \mathbb{V}_k^{M_2}} \rho(\mathbf{E}_{w}^{M_2})
%         \right)\,,
% \end{split}
% % \begin{split}
% %  \{\mathbf{M}\}= \underset{\mathbf{M}} {\operatorname{argmin}}\sum_{k \in \mathcal{K_c}}\left[\sum_{v \in \mathbb{V}_k}  \right.\\ \left.  \left( \rho\left(\mathbf{E}^{\parallel}{_{k,v,M_0}}\right) + \rho\left(\mathbf{E}^{\perp}{_{k,v,M_1}} \right) + \rho\left(\mathbf{E}^{\perp}{_{k,v,M_2}} \right)\right)\right]\,,
% % \end{split}
% \end{equation}
\begin{equation}
\begin{split}
    \mathcal{M} = \underset{\mathcal{M}}{\operatorname{argmin}} 
        \sum_{k \in \mathcal{K}_M}
        \left(
            \sum_{j \in \mathbb{V}_k^{\text{MA}_0}} \rho(\mathbf{E}_{j}^{\text{MA}_0})
          + \sum_{j \in \mathbb{V}_k^{\text{MA}_1}} \rho(\mathbf{E}_{j}^{\text{MA}_1}) 
        \right.
        & \\
        \left.
          + \sum_{j \in \mathbb{V}_k^{\text{MA}_2}} \rho(\mathbf{E}_{j}^{\text{MA}_2})
        \right)\,,
\end{split}
\end{equation}
% where the error term of a line $i$ associated to the axis $M_j$ is defined as:
where the error term of a line $j$ associated to the axis $M_j \in \mathcal{M}$ is given by:
% \begin{equation}
% \mathbf{E}_{i}^{M_j} = \sqrt{1 - d^{\,2}(L_i^w, M_j)} + d(L_i^w, M_m) + d(L_i^w, M_n) \cdot \omega_i^{-1}\,,
% %\mathbf{E}^{\parallel}_{k,v,M_0} = \left\|1 -\left( \frac{\left( (R_{k}E_v + t_k) - (R_{k}S_v + t_k) \right) \cdot M_0}{\left\|(R_{k}E_v + t_k) - (R_{k}S_v + t_k)\right\| \left\|  M_0\right\|} \right)\right\|_{\Sigma}^{2}.
% \end{equation}
% \begin{equation}
% \mathbf{E}_{i}^{M_j} = \left[\sqrt{1 - d^{\,2}(L_i^w, M_j)} + d(L_i^w, M_m) + d(L_i^w, M_n)\right] \cdot \omega_i^{-1}\,,
% \end{equation}
\begin{equation}
\mathbf{E}_{j}^{M_j} = \mathbf{E}_{j,M_j}^{\parallel} + 
                       \mathbf{E}_{j,M_{j\prime}}^{\perp} +
                       \mathbf{E}_{j,M_{j\prime\prime}}^{\perp} \,,
\end{equation}
% being $M_{j\prime}$ and $M_{j\prime\prime}$ the MA non-associated to the line. To conclude, we apply the Single Value Decomposition (SVD) to preserve the orthogonality between axes. %%To conclude, the proposed coarse to fine strategy allows using a new set of features to extract the MA, where the traditional Mean Shift algorithm produce inaccuracies.
being $M_{j\prime}$ and $M_{j\prime\prime}$ the two other MA non-associated to line $j$. To conclude, we apply Single Value Decomposition (SVD) to ensure the orthogonality between axes.

% where, the parallel error term related is denoted as:  
% \begin{equation}
% \label{eq-manh_axis_par}
% \mathbf{E}^{\parallel}_{k,v,M_0} = \left\|1 -\left( \frac{\left( (R_{k}E_v + t_k) - (R_{k}S_v + t_k) \right) \cdot M_0}{\left\|(R_{k}E_v + t_k) - (R_{k}S_v + t_k)\right\| \left\|  M_0\right\|} \right)\right\|_{\Sigma}^{2}.
% \end{equation}

%  The perpendicular angle error of a map line observation with its non-associated axes $H \in (M_1 \cup M_2)$ is given by:  
% \begin{equation}
% \mathbf{E}^{\perp}_{k,v,H} = \left\| \frac{\left( (R_{k}E_v + t_k) - (R_{k}S_v + t_k) \right) \cdot H}{\left\|(R_{k}E_v + t_k) - (R_{k}S_v + t_k)\right\| \left\|  H\right\|} \right\|_{\Sigma}^{2}.
% \end{equation}

\section{Experimental Results}
\label{sec:results}
%To demonstrate the performance of the proposed approach, we conduct various experiments in both synthetic and real image sequences. Additionally, we compare its localization accuracy with some state-of-the-art VO and visual SLAM systems. The datasets used for the evaluation are: the ICL-NUIM dataset~\cite{handa2014ICL} and the TUM RGB-D benchmark~\cite{sturm2012TUMDataset}. 
To demonstrate the performance of \approachname{}, we conduct various experiments in both synthetic and real image sequences. Additionally, we compare its localization accuracy with some state-of-the-art VO and visual SLAM systems by means of the following datasets: 
\begin{itemize}
    % \item \textbf{ICL-NUIM Dataset}: The ICL-NUIM is a synthetic dataset which comprises two main scenes, the living room and the office, coined in our experiments as $lr$ and $of$ respectively. Furthermore, this is an indoor dataset with large structure areas, where the MW assumption and the structural constraints are highly present. Additionally, this dataset involves some low-textured challenging scenes such as floors, ceilings and walls. 
    % \item \textbf{TUM RGB-D Benchmark} The TUM RGB-D benchmark is also an indoor dataset that contains several sequences with different structure, illumination and texture conditions. Unlike ICL-NUIM, this dataset contains some sensor noise because of a real RGB-D sensor is used.   
    \item \textbf{ICL-NUIM Dataset}~\cite{handa2014ICL}. This is a synthetic dataset which comprises two main scenes, the living room and the office, coined in our experiments as $lr$ and $of$, respectively. Furthermore, this is an indoor dataset with large structured areas, where the MW assumption and the structural constraints are highly present. Additionally, this dataset involves some low-textured challenging elements such as floors, ceilings and walls. 
    \item \textbf{TUM RGB-D Benchmark}~\cite{sturm2012TUMDataset}. This is also an indoor dataset that contains several sequences with different structure, illumination and texture conditions. Unlike ICL-NUIM, this dataset contains some sensor noise because a real RGB-D sensor is used.   
\end{itemize}

% Regarding \approachname{}{} parameters, we have used the default parameters provided by ORB-SLAM2 in those common parts. Whereas, the parameter values of the rest are obtained experimentally for a single dataset, and they remain fixed for the rest of sequences.
Regarding the \approachname{}{} parameters, we have used the default values provided by ORB-SLAM2 authors for the common parts, whereas the remaining parameters have been set experimentally from a single dataset, and they have been kept unaltered for the rest of sequences.

% To evaluate the overall performance of the proposed solution, we use the Root-Mean-Square Error (RMSE) of the Absolute Position Error (APE) expressed in meters. The localization accuracy, as well as the trajectory plots, have been extracted using the EVO library~\cite{grupp2017evo}. All the conducted experiments have been performed on an Intel Core i7-9750H CPU @ 2.60GHz and 16GB RAM without GPU parallelization.
To evaluate the overall performance of \approachname{}, we use the Root-Mean-Square Error (RMSE) of the Absolute Position Error (APE) expressed in meters. The localization accuracy, as well as the trajectory plots, have been generated using the EVO library~\cite{grupp2017evo}. All the experiments have been performed on an Intel Core i7-9750H~@~2.60GHz~/~16GB RAM, without GPU parallelization.

\subsection{General Performance}
\begin{figure}[tb]
    \centering
    % \begin{tabular}{@{\hspace{0mm}}c@{\hspace{1mm}}c@{\hspace{0mm}}}
    % \includegraphics[width=0.48\linewidth, clip=true, trim=0 20 40 0]{figures/no_MA} &
    % \includegraphics[width=0.48\linewidth, clip=true, trim=0 35 655 60]{figures/fr3-longoffice} \\
    % \end{tabular}
    \includegraphics[align=c,width=0.49\linewidth, clip=true, trim=0 20 40 0]{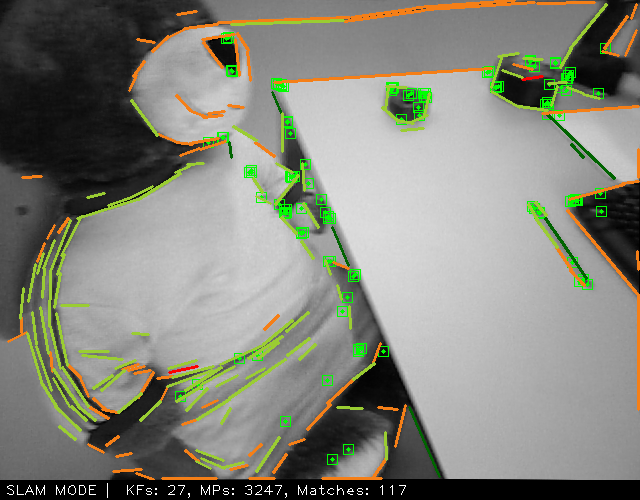}
    %\qquad
    \includegraphics[align=c,width=0.47\linewidth, clip=true, trim=0 35 655 60]{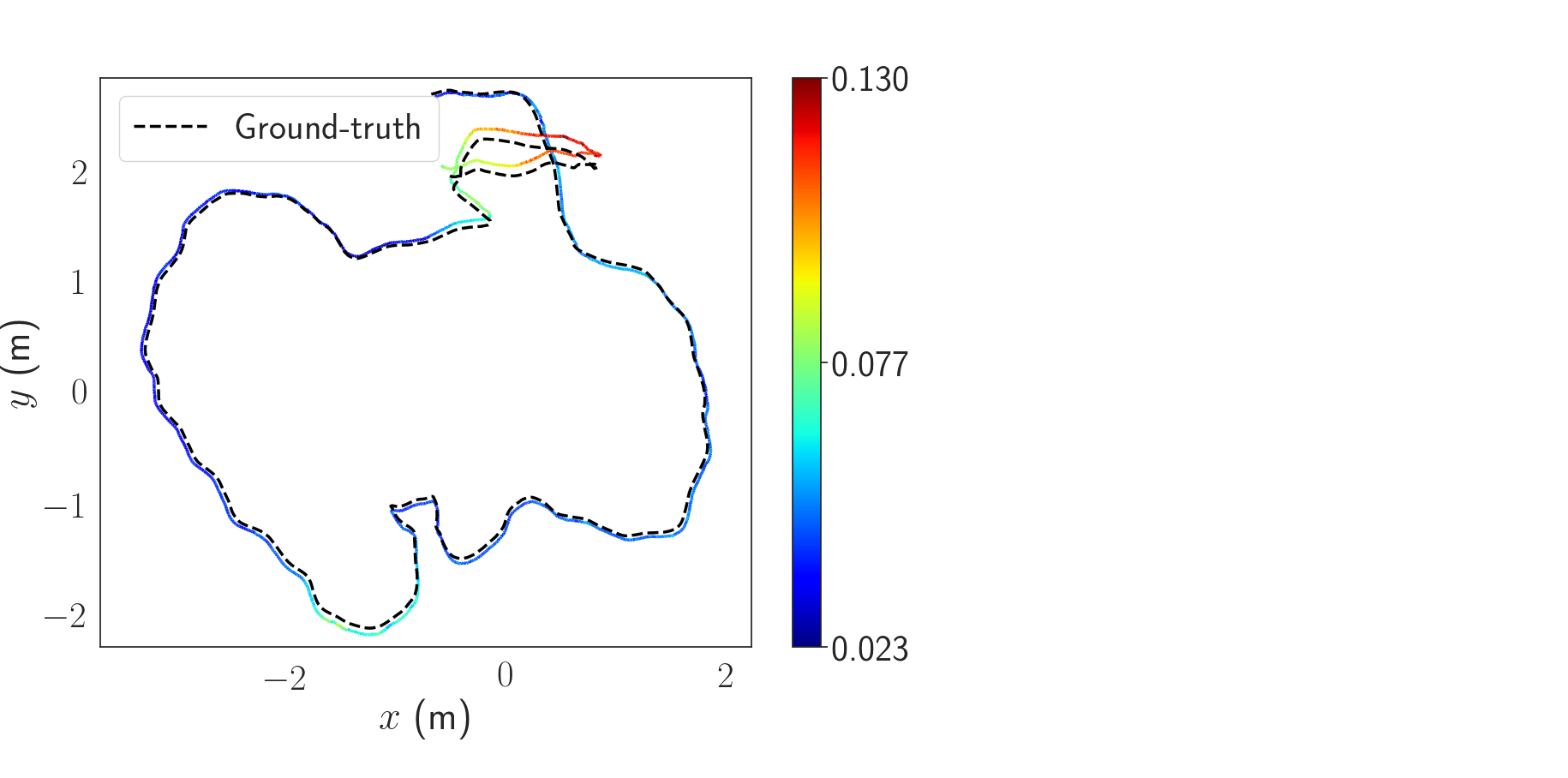}
%  \vspace{-4.5mm}
    %\caption{Left-image illustrates the absence of the MA in a scene of the fr3-longoffice dataset. Right-image depicts the trajectory of the MSC-VO in this dataset, where no tracking failures are observed.}
    \caption{(left) The MA maybe absent in a scene, e.g. a frame of the \emph{fr3-longoffice} sequence. (right) Trajectory estimated by \approachname{} for this sequence, where no tracking failures are observed.}
    \label{fig:failure_n_trajectory}
\end{figure}

% As stated previously, the use of the MA in a local map optimization avoids that the absence of the MA produces a tracking failure. It affects to those works that rely on the MA as the main source of the tracking. This effect is illustrated on the left-image of Figure \ref{fig:failure_n_trajectory}. This image corresponds to a frame of the fr3-longoffice dataset, where the MW assumption is not accomplished. In this figure, green, red and blue colour represent the correspondence of a line with a single Manhattan Axis. Whereas, yellow shows that the 3D line representation do not correspond to an axis, and orange represent that the 3D line position estimation has not been computed. Moreover, the right-image demonstrate that the presented work is able to compute the whole trajectory. As can be observed in the last part of this trajectory, which corresponds to the area with the highest error, the rotational error is almost negligible.
For a start, Fig.~\ref{fig:failure_n_trajectory} illustrates the fact that the MA may be absent in a scene, leading to tracking failures for some solutions. In the case of \approachname{}, the fact of involving the MA only in local map optimizations can prevent these failures from occurring. In Fig.~\ref{fig:failure_n_trajectory}(left), we show a frame of the \emph{fr3-longoffice} sequence, for which the MW assumption is not very appropriate. In the image, green, red and blue colours denote the correspondences of a line with a single Manhattan axis, whereas yellow is for 3D lines that do not correspond to any axis and orange is for lines whose 3D position has not been estimated. Figure~\ref{fig:failure_n_trajectory}(right) shows that \approachname{} can estimate the whole trajectory. The highest estimation errors can be observed at the end of the trajectory, and mostly on the translation component of motion.

% In the following, we assess the net effect of the proposed contributions to the \approachname{} performance. To this end, we differentiate three approaches. The first one corresponds to a VO which combines point and line features (PL-VO). The second one includes the depth extraction procedure on top of the PL-VO (PL-VO-Depth). Whereas, the third one corresponds to the proposed work (\approachname{}). Table \ref{tab:contributions} present the localization accuracy of these methods in multiple datasets. As can be observed in this table, every contribution of the proposed work increase notably the localization accuracy in each dataset. 
Next, we compare several versions of \approachname{} to show the effect of the different contributions: \emph{PL-VO} is the part of \approachname{} that just combines point and line features, \emph{PL-VO-Depth} combines PL-VO with depth calculations, and the third case is the full version of \approachname{}. Estimation performance results for multiple sequences can be found in Table~\ref{tab:contributions}. As can be observed, the estimation accuracy improves notably for all sequences considered as we incorporate more features in the VO. 

\begin{table}[tb]
\centering
% \caption{RMSE of the APE (m) of the proposed approach, with the main contributions of this work evaluated in multiple sequences.}
\caption{RMSE of the APE of \approachname{} (in meters)}
\vspace{-2mm}
\begin{tabular}{|c|c|c|c|}
\hline
Sequence & PL-VO & PL-VO-Depth & \approachname{} \\ \hline
lr-kt0        & 0.051      & 0.024            & 0.010 \\ \hline
lr-kt1        & 0.064      & 0.048            & 0.015 \\ \hline
lr-kt2        & 0.054      &  0.030           & 0.022 \\ \hline
lr-kt3        & 0.061      & 0.057            & 0.041 \\ \hline
of-kt0        & 0.047      & 0.032            & 0.029 \\ \hline
of-kt1        & 0.056      & 0.053            & 0.030 \\ \hline
of-kt2        & 0.040      & 0.039            & 0.025 \\ \hline
of-kt3        & 0.042      &  0.038           & 0.022 \\ \hline
% snot-far      &       &             & 0.066 \\ \hline
% snot-near     &       &             & 0.00 \\ \hline
large-cabinet & 0.173      & 0.152            & 0.135 \\ \hline
fr3-longoffice& 0.108      &  0.096           & 0.049 \\ \hline
\end{tabular}
% \vspace{-0.65cm}
\label{tab:contributions}
\end{table}

\begin{figure*}[tb]
    \centering
    \begin{tabular}{@{\hspace{0mm}}c@{\hspace{1mm}}c@{\hspace{1mm}}c@{\hspace{1mm}}c@{\hspace{0mm}}}
    \includegraphics[ width=0.24\linewidth, clip=true, trim= 200 85 140 80 ]{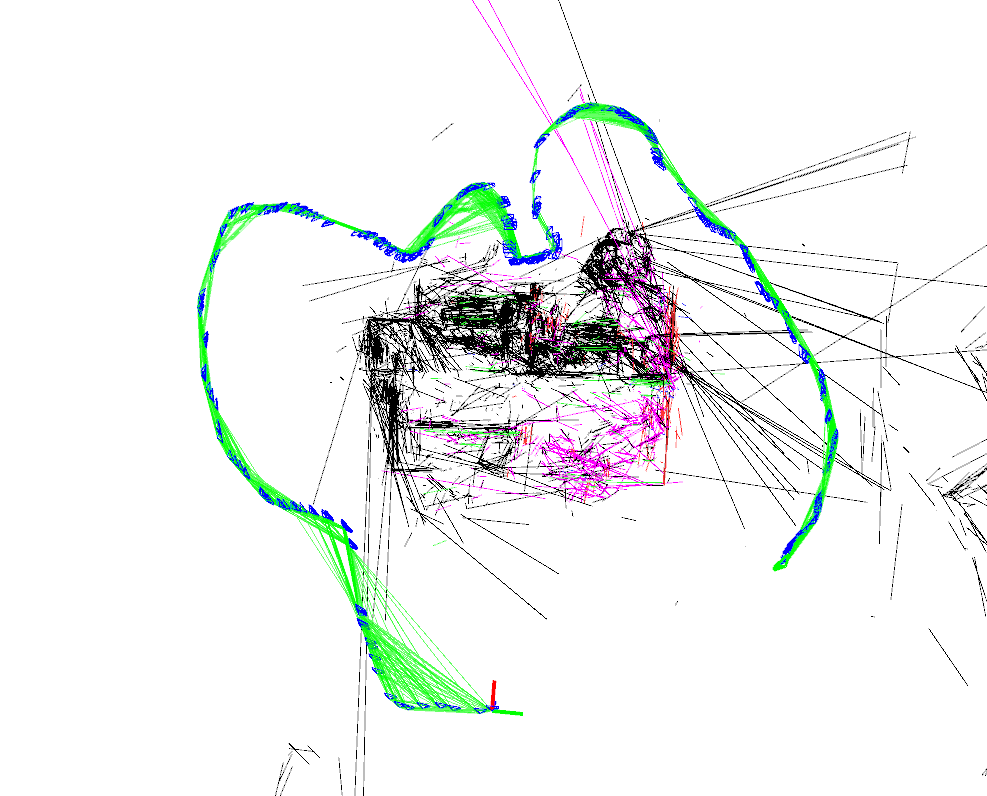} &
    \includegraphics[width=0.24\linewidth, clip=true, trim= 50 50 120 80 ]{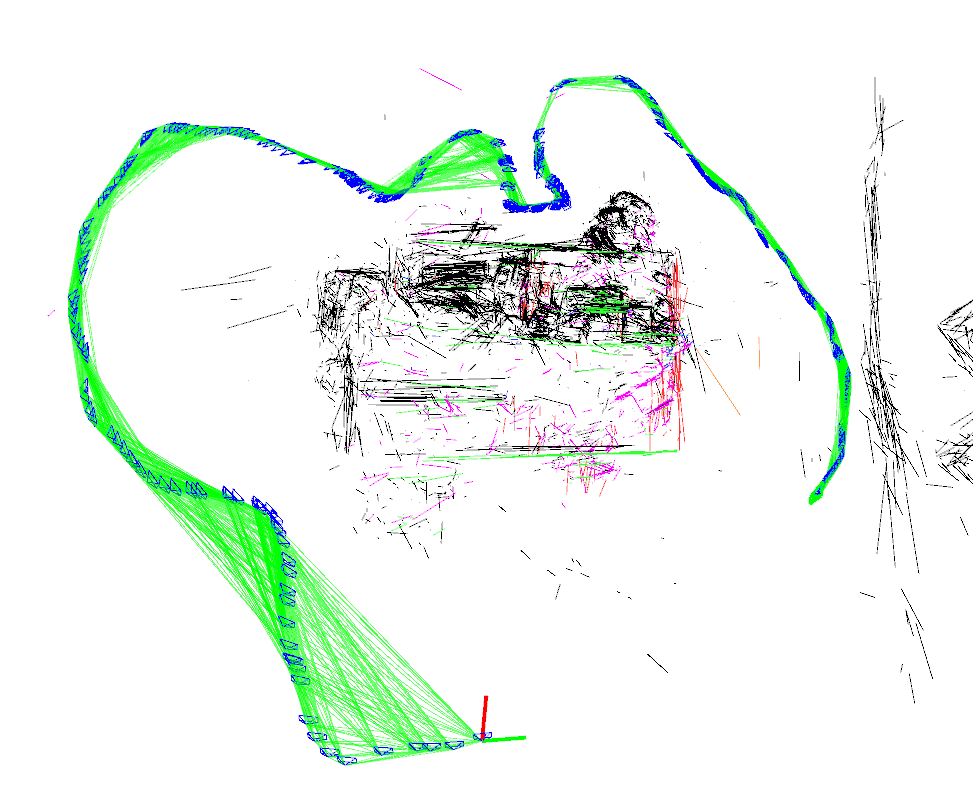} &
    \includegraphics[width=0.24\linewidth, clip=true, trim= 50 35 160 60]{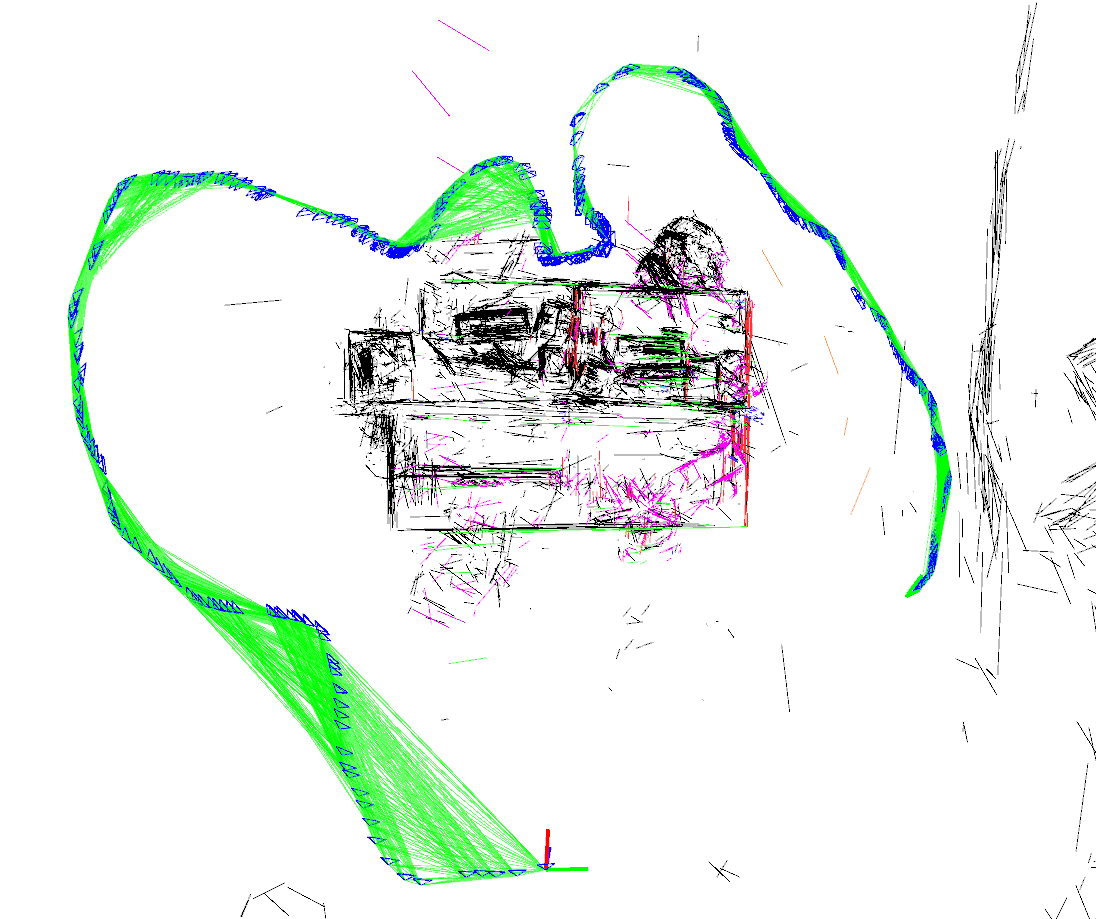} & 
    \includegraphics[width=0.24\linewidth, clip=true, trim= 180 20 580 0]{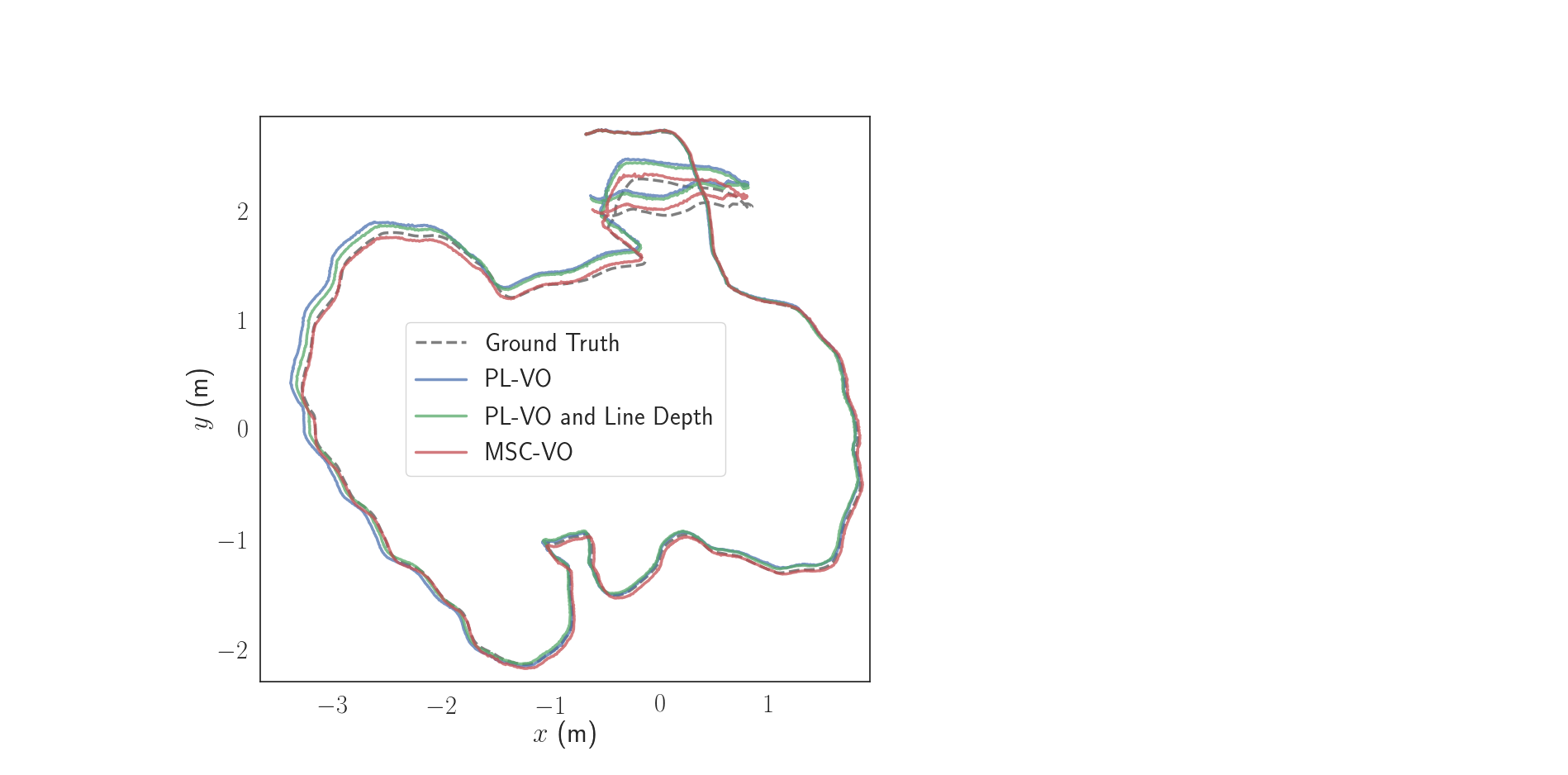} 
    \end{tabular}
%  \vspace{-4.5mm}
    % \caption{The three left images illustrate the resulting local maps by comparing different contributions of this work in the fr3-longoffice sequence. The first image corresponds to the resulting local map using a point and line VO (PL-VO). The second image represent the PL-VO using the proposed line depth extraction procedure, while the third image corresponds to the \approachname{}. The fourth image illustrates the 2D trajectories of each presented approach. These are represented in blue, green and red, respectively. Additionally, the ground truth is depicted using the dashed grey colour.}
    \caption{(left) Local maps for the \emph{fr3-longoffice} sequence and different versions of \approachname{}: 1st --  only using points and lines (PL-VO), 2nd -- PL-VO using the proposed line depth extraction procedure (PL-VO-Depth), 3rd -- full \approachname{}. (right) 2D trajectories for PL-VO, PL-VO-Depth and \approachname{}, respectively shown in blue, green and red, and the ground truth in dashed grey.}
    \label{fig:trajectories}
\end{figure*}

% To illustrate the effect of the proposed contributions, Figure \ref{fig:trajectories} shows the resulting local map of the fr3-longoffice dataset for the three above-mentioned approaches. The first image illustrates the result of the PL-VO and shows how the noise of the lines affect the local map, and consequently, the accuracy of the pose estimation decreases. The second image represents the PL-VO-Depth. In this image, the noise of the line depth extraction procedure is reduced. However, there are still pose inaccuracies. The third image shows the result of the \approachname{}, where, as can be observed, present the best line representations in the map, and the best localization accuracy of these three approaches. Point out that, the local map optimization procedure not only improves the camera pose accuracy, but also refines the map lines from the local map. As a result, the misalignment shown in the PL-VO-Depth method is notably reduced. Note that, for better reader comprehension, the performed approaches of the three first figures have been stopped before finishing. To conclude, for a further understanding of the achieved pose accuracy by each approach, the fourth image plots the trajectories regarding each method against the ground truth.
On the other side, Fig.~\ref{fig:trajectories} shows local maps from the same cases as above for the \emph{fr3-longoffice} sequence. The first and second plots result from, respectively, PL-VO and PL-VO-Depth. In the former case, noise from lines depth calculation affects the local map and, consequently, also the pose estimation accuracy. In the second case, this noise is of a lower magnitude, but pose inaccuracies are still observed. The third plot results from \approachname{} with the best local map and the highest localization accuracy. These results show that the local map optimization procedure not only improves the camera pose accuracy, but also refines the map lines. As a result, the misalignment that affects the PL-VO-Depth case is notably reduced. To conclude, the fourth plot shows the trajectories from each approach together with the ground truth, for a further understanding of the pose accuracy achievable on each case.

% Finally, the computation times of \approachname{} are presented in Table \ref{tab:comp_times}. This table summarizes the execution times of the main stages presented in this work. These times are extracted using the mean computational times of three different sequences from the TUM RBG-D benchmark. As a general overview, it is well known that adding line features into point based VO or SLAM methods improves the accuracy and the robustness, but it also increases the required computational times~\cite{Pumarola2017PLSLAM}. In more detail, despite the low-time required to extract point and line features, it is penalized by the robust fitting method used in the 3D line pose estimation. Regarding the coarse MA estimation procedure, its execution time is high compared to the rest of the process of the tracking. However, in scenes where the MW assumption adhere, it is only computed once. Despite the local map optimization requires more time than the traditional methods based on local bundle adjustment, this is fast enough for our purpose, as it runs in a parallel thread. To sum up, the final frame rate of the proposed method is around 18 Hz.
% CONTRASTARLO ---> which is a challenging time compared to the state-of-the-art works.
To finish, average running times for the main stages of \approachname{} can be found in Table~\ref{tab:comp_times}. The averages result from three different sequences of the TUM RBG-D benchmark. As expected, adding line features into point based VO or SLAM methods improves the accuracy and the robustness, though at the expense of increasing the computational complexity~\cite{Pumarola2017PLSLAM}. In more detail regarding our solution: (1) the robust fitting method used for 3D line pose estimation increases the low times required to extract line features and adds execution time to the feature extraction stage over other solutions; (2) regarding MA estimation, its execution time is high due to 180.4 ms that are required by the coarsest estimation step, although it needs to be computed only once (in scenarios where the MW assumption holds); and (3) despite local map optimizations require more time than other, more traditional methods based on local bundle adjustment, it can still be fast enough, as they run in a parallel thread. As a general comment, the final frame rate achieved is around 18 Hz.

\begin{table}[tb]
\centering
% \caption{Mean execution time for three sequences of the TUM RGB-D benchmark. These times are computed for the main processes explained in this work. As an additional remark, the coarse and fine MA estimation methods are processed only once during the whole dataset.}
\caption{Mean execution times (TUM RGB-D benchmark)}
\vspace{-2mm}
\begin{tabular}{|c|c|c|c|c|}
\hline
\multicolumn{5}{|c|}{Mean Execution Time (ms)}\\ \hline
\multicolumn{3}{|c|}{Tracking} & \multicolumn{2}{c|}{Local Mapping} \\ \hline
\begin{tabular}[c]{@{}c@{}}Feat. Extrac.\\  and 3D Pose \\ Estim.\end{tabular} & \begin{tabular}[c]{@{}c@{}}Camera \\ Pose \\ Estim.\end{tabular} &  \begin{tabular}[c]{@{}c@{}}Total\\ (Hz)\end{tabular} & \begin{tabular}[c]{@{}c@{}}Local Map \\ Optim.\end{tabular} & \begin{tabular}[c]{@{}c@{}}MA\\ Estim.\end{tabular} \\ \hline
23.2 & 29.1 & 18 & 152.6 & 206.6 \\ \hline
\end{tabular}
% \vspace{-0.65cm}
\label{tab:comp_times}
\end{table}

%%% AOR %%%

\subsection{Comparison with other solutions}
% In this section, we compare the localization accuracy with multiple state-of-the-art approaches. Table \ref{tab:rmse_transl}, shows the localization accuracy of \approachname{} compared to the related works. he results reported come from the original works. The better localization accuracy is indicated in bold face, whereas the second one is illustrated in blue. n.a. indicates that this value is not available, whereas, $X$ indicates a tracking failure. In the left-side of the table, we show the accuracy of three VO that benefits from the MA assumption ~\cite{zhou2016divide,kim2017OPVO,kim2018LPVO}. As can be observed in this part, our approach achieves the best accuracy, in all the datasets except the snot-near, where a low number of feature detection produces a tracking failure. Note that, this effect is not present in those works that rely on planar features due to the continuous existence of coplanar planes of the scene. Conversely, the right-side of the table, shows the results obtained by multiple SLAM approaches, where the proposed work compares favourably.   
Table~\ref{tab:rmse_transl} compares \approachname{} regarding localization accuracy with other state-of-the-art approaches, for which the results reported in the original works are reproduced. Best performances are indicated in bold face, whereas the second best is shown in blue, \emph{n.a.} refers to a not-available value, and $\times$ reports a tracking failure. The left group of the table comprises \approachname{} and three VO that also benefit from the MA assumption. As can be observed, among the VO considered, \approachname{} achieves the highest accuracy for all datasets except for \emph{snot-near}, where a tracking failure takes place. This effect is not present in those works that rely on planar features due to the continuous presence of coplanar planes in the different scenes. The right group of the table reports on the results obtained by several SLAM approaches, with which \approachname{} compares favourably even without loop closing, as can be observed.

% Among these works, the most similar to ours is presented in~\cite{Li2021RGB-DSLAM}, where two solutions are presented. The first one, coined as SReg-wo, uses the MA during the tracking procedure. While the second, SReg, extends the previous procedure by using a later refinement step which combine parallel and perpendicular constraints of planes.   

\begin{table*}
\centering
% \caption{RMSE of the APE (m) of the proposed approach, compared with multiple state-of-the-art VO and SLAM approaches in multiple sequences. $\times$ represents a tracking failure, where n.a. indicates that this value is not available. The lowest error is indicated in bold face, while the second one is indicated in blue.}
\caption{RMSE of the APE for \approachname{} and other state-of-the-art VO and SLAM approaches (in meters)}
\vspace{-2mm}
\resizebox{\textwidth}{!}{%
\begin{tabular}{c|cccc|cccccc}\hline
\multicolumn{1}{c|}{} & \multicolumn{4}{c|}{Visual Odometers} & \multicolumn{6}{c}{Visual SLAM Approaches} \\ \hline 
%\multicolumn{1}{c|}{Sequence}& \multicolumn{1}{c}{MSC-VO} & \multicolumn{1}{c}{OPVO~\cite{kim2017OPVO}}& 
\multicolumn{1}{c|}{Sequence}& \multicolumn{1}{c}{\approachname{}} & \multicolumn{1}{c}{OPVO~\cite{kim2017OPVO}}& 
\multicolumn{1}{c}{LPVO~\cite{kim2018LPVO}} & \multicolumn{1}{c|}{MWO~\cite{zhou2016divide}} & \multicolumn{1}{c}{SReg-wo~\cite{Li2021RGB-DSLAM}} & \multicolumn{1}{c}{SReg~\cite{Li2021RGB-DSLAM}} & \multicolumn{1}{c}{ORB-SLAM2~\cite{Mur2017ORBSLAM2}} & \multicolumn{1}{c}{PS-SLAM~\cite{Zhang2019PointPLane}} &  \multicolumn{1}{c}{L-SLAM~\cite{kim2018linear}}  & \multicolumn{1}{c}{InfiniTAM~\cite{prisacariu2017infinitam}} \\ \hline
lr-kt0    & \color{blue}{\textbf{0.010}} & $\times$   &0.015 &$\times$  &0.025 &\textbf{0.006}    & 0.025 & 0.016   & 0.012     & $\times$ \\
lr-kt1    &  0.015 & 0.04 & 0.039  &0.32    &0.036 & 0.015         & \color{blue}{\textbf{0.008}} & 0.018   & 0.027    & \textbf{0.006} \\
lr-kt2   &  0.022 & 0.06 & 0.034  &0.11 &0.053 & 0.020   & 0.023 & \color{blue}{\textbf{0.017}}   & 0.053    & \textbf{0.013}  \\
lr-kt3    &  0.041  & 0.10 & 0.102  &0.40  &0.059 &\textbf{0.012}     & \color{blue}{\textbf{0.021}} & 0.025   & 0.143     & $\times$       \\  \hline
of-kt0    &  \color{blue}{\textbf{0.029}}  &0.06 & 0.061  &0.31  &0.068 & 0.041        & 0.037 & 0.032   & \textbf{0.020}  & 0.042 \\
of-kt1    &  0.030 &0.05 & 0.052  &1.10  &0.028 & 0.020  & 0.029 & \color{blue}{\textbf{0.019}}   & \textbf{0.015}  & 0.025   \\
of-kt2   &  \color{blue}{\textbf{0.025}}  &$\times$ & 0.039  &$\times$  &0.06 & \textbf{0.011}     & 0.039 & 0.026   & 0.026     &$\times$ \\
of-kt3    &  0.022 & 0.04  & 0.030  &1.38  &0.012 & 0.014         & 0.065 & 0.012  & \color{blue}{\textbf{0.011}}  & \textbf{0.010} \\  \hline
snot-far &  0.066 &0.13 & 0.075  &0.47  &0.026  & \color{blue}{\textbf{0.022}}   & $\times$ & \textbf{0.020} & 0.141    & 0.037 \\
snot-near   & $\times$ &0.16   & 0.080  &0.95 &$\times$  & 0.025   & $\times$ & \textbf{0.013} & 0.066  & \color{blue}{\textbf{0.022}} \\
% % cabinet* & Add   & 0.333 &n.a. & 0.520  &n.a. &n.a. &0.057 & \textbf{0.035}     & 0.075 & 0.067   & 0.291    & 0.690 & 0.035   \\
large-cabinet   & 0.135  &0.51  & 0.279  &0.83 &0.813 & \textbf{0.071}  & 0.124 & \color{blue}{\textbf{0.079}}   & 0.140   & 0.512    \\
fr3-longoffice & \color{blue}{\textbf{0.049}} &$\times$  & 0.19  &$\times$  &n.a. &n.a.   & \textbf{0.02} & n.a.   & n.a.  & n.a.    \\\hline 
\multicolumn{11}{l}{} \\[-2mm]
\multicolumn{11}{l}{\small $\times$ and n.a. respectively stand for \emph{tracking failure} and \emph{not available} value. The best result for each sequence is shown in bold and the second best in blue.}
\end{tabular}
}
% \vspace{-0.65cm}
\label{tab:rmse_transl}
\end{table*}

%%%%%%%%%%%%%%%%%%%%%%%%%%%%%%%%%%%%%%%%%%%%%%%%%%%%%%%%%%%%%%%%%%%%%%%%%%%%%%%%%%%%%
\section{Conclusions and future work}
\label{sec:conclusions}
% In this work, we have described \approachname{} a VO that improves the camera pose accuracy in human-made environments. This is achieved by a point and line VO that leverages, if exist, the structural regularities and the MW assumption of the environment. On one side, the structural constraints are used to improve the line depth extraction, and the MA estimation procedure. On the other side, these structural constraints are combined with the point and line reprojection error and the MW assumption in a local map optimization procedure. All these contributions allow to \approachname{} to increase the accuracy of the 3D map lines position and the computed trajectory. Furthermore, unlike those state-of-the-art works that use the MW in the tracking stage, our pipeline is designed to deal with the absence of the MA, allowing to work in a wider range of environments.   
In this work, we have described \approachname{}, a VO that improves camera pose estimation accuracy in human-made environments. This is achieved by a combined point and line VO approach that leverages the structural regularities of the environment as well as the satisfaction of the MW assumption. On the one side, the structural constraints are used to improve line depth extraction and MA estimation. On the other side, these structural constraints are combined with point and line reprojection errors together with the MW assumption for local map optimization. All these contributions have been shown to increase the accuracy of 3D map lines position estimation and the computed trajectory for \approachname{}. Furthermore, contrary to other state-of-the-art works that use the MW in the tracking stage, our pipeline is designed to deal with the absence of the MA, allowing us to work in a wider range of environments.   

% Regarding future work, we plan to integrate an incremental loop closure detection to the proposed VO. We also plan to introduce the structural constraints and the MA alignment into a global map optimization.  
Regarding future work, we plan to integrate \approachname{} with an incremental loop closure detection strategy. We are also intent to make use of the structural constraints and the MA alignment for global map optimization.

\bibliographystyle{IEEEtran}
\bibliography{JPC}

\end{document}